%% file: main.tex
\documentclass[10pt]{article}

\usepackage[T1]{fontenc}
\usepackage{microtype}
\usepackage{amsmath}
\usepackage{amssymb}
\usepackage{tikz}
\usetikzlibrary{arrows.meta,positioning,fit,backgrounds,calc}
\usepackage{booktabs}
\usepackage{graphicx}
\usepackage{float}
\usepackage{placeins}
\usepackage[letterpaper]{geometry}
\usepackage{natbib}
\usepackage[hidelinks]{hyperref}

\makeatletter

\renewcommand{\normalsize}{%
  \@setfontsize\normalsize\@xpt\@xipt
  \abovedisplayskip      7\p@ \@plus 2\p@ \@minus 5\p@
  \abovedisplayshortskip \z@ \@plus 3\p@
  \belowdisplayskip      \abovedisplayskip
  \belowdisplayshortskip 4\p@ \@plus 3\p@ \@minus 3\p@
}
\normalsize
\renewcommand{\small}{%
  \@setfontsize\small\@ixpt\@xpt
  \abovedisplayskip      6\p@ \@plus 1.5\p@ \@minus 4\p@
  \abovedisplayshortskip \z@  \@plus 2\p@
  \belowdisplayskip      \abovedisplayskip
  \belowdisplayshortskip 3\p@ \@plus 2\p@   \@minus 2\p@
}
\renewcommand{\footnotesize}{\@setfontsize\footnotesize\@ixpt\@xpt}
\renewcommand{\scriptsize}{\@setfontsize\scriptsize\@viipt\@viiipt}
\renewcommand{\tiny}{\@setfontsize\tiny\@vipt\@viipt}
\renewcommand{\large}{\@setfontsize\large\@xiipt{14}}
\renewcommand{\Large}{\@setfontsize\Large\@xivpt{16}}
\renewcommand{\LARGE}{\@setfontsize\LARGE\@xviipt{20}}
\renewcommand{\huge}{\@setfontsize\huge\@xxpt{23}}
\renewcommand{\Huge}{\@setfontsize\Huge\@xxvpt{28}}

\renewcommand{\section}{%
  \@startsection{section}{1}{\z@}%
                {-2.0ex \@plus -0.5ex \@minus -0.2ex}%
                { 1.5ex \@plus  0.3ex \@minus  0.2ex}%
                {\large\bf\raggedright}%
}
\renewcommand{\subsection}{%
  \@startsection{subsection}{2}{\z@}%
                {-1.8ex \@plus -0.5ex \@minus -0.2ex}%
                { 0.8ex \@plus  0.2ex}%
                {\normalsize\bf\raggedright}%
}
\renewcommand{\subsubsection}{%
  \@startsection{subsubsection}{3}{\z@}%
                {-1.5ex \@plus -0.5ex \@minus -0.2ex}%
                { 0.5ex \@plus  0.2ex}%
                {\normalsize\bf\raggedright}%
}
\renewcommand{\paragraph}{%
  \@startsection{paragraph}{4}{\z@}%
                {1.5ex \@plus 0.5ex \@minus 0.2ex}%
                {-1em}%
                {\normalsize\bf}%
}
\renewcommand{\subparagraph}{%
  \@startsection{subparagraph}{5}{\z@}%
                {1.5ex \@plus 0.5ex \@minus 0.2ex}%
                {-1em}%
                {\normalsize\bf}%
}

\newlength{\@nipsabovecaptionskip}
\newlength{\@nipsbelowcaptionskip}
\renewenvironment{table}
  {\setlength{\abovecaptionskip}{\@nipsbelowcaptionskip}%
   \setlength{\belowcaptionskip}{\@nipsabovecaptionskip}%
   \@float{table}}
  {\end@float}

\renewcommand{\footnoterule}{\kern-3\p@ \hrule width 12pc \kern 2.6\p@}

\def\@listi{\leftmargin\leftmargini}
\def\@listii{\leftmargin\leftmarginii
              \labelwidth\leftmarginii
              \advance\labelwidth-\labelsep
              \topsep  2\p@ \@plus 1\p@    \@minus 0.5\p@
              \parsep  1\p@ \@plus 0.5\p@ \@minus 0.5\p@
              \itemsep \parsep}
\def\@listiii{\leftmargin\leftmarginiii
              \labelwidth\leftmarginiii
              \advance\labelwidth-\labelsep
              \topsep    1\p@ \@plus 0.5\p@ \@minus 0.5\p@
              \parsep    \z@
              \partopsep 0.5\p@ \@plus 0\p@ \@minus 0.5\p@
              \itemsep \topsep}
\def\@listiv{\leftmargin\leftmarginiv
              \labelwidth\leftmarginiv
              \advance\labelwidth-\labelsep}
\def\@listv{\leftmargin\leftmarginv
              \labelwidth\leftmarginv
              \advance\labelwidth-\labelsep}
\def\@listvi{\leftmargin\leftmarginvi
              \labelwidth\leftmarginvi
              \advance\labelwidth-\labelsep}

\newcommand{\@toptitlebar}{%
  \hrule height 4\p@
  \vskip 0.25in
  \vskip -\parskip
}
\newcommand{\@bottomtitlebar}{%
  \vskip 0.29in
  \vskip -\parskip
  \hrule height 1\p@
  \vskip 0.09in
}
\renewcommand{\maketitle}{%
  \par
  \begingroup
    \renewcommand{\thefootnote}{\fnsymbol{footnote}}
    \renewcommand{\@makefnmark}{\hbox to \z@{$^{\@thefnmark}$\hss}}
    \long\def\@makefntext##1{%
      \parindent 1em\noindent
      \hbox to 1.8em{\hss $\m@th ^{\@thefnmark}$}##1
    }
    \thispagestyle{empty}
    \@maketitle
    \@thanks
  \endgroup
  \let\maketitle\relax
  \let\thanks\relax
}
\renewcommand{\@maketitle}{%
  \vbox{%
    \hsize\textwidth
    \linewidth\hsize
    \vskip 0.1in
    \@toptitlebar
    \centering
    {\LARGE\bf \@title\par}
    \@bottomtitlebar
    \ifx\@author\@empty
    \else
      \begin{tabular}[t]{c}\bf\rule{\z@}{24\p@}\@author\end{tabular}%
      \vskip 0.3in \@minus 0.1in
    \fi
  }
}

\renewenvironment{abstract}%
{%
  \vskip 0.075in
  \centerline{\large\bf Abstract}
  \vspace{0.5ex}
  \begin{quote}
}
{
  \par
  \end{quote}
  \vskip 1ex
}

\makeatother

\title{Uniform Herding: Exemplar Replay with Representation Refresh}
\author{Krishna Subedi \\ \texttt{krishna.subedi@neryva.com}}

\begin{document}

\maketitle

\begin{abstract}
  As the feature representation changes, replay must preserve the earlier classes. However, only a bounded active exemplar set can be replayed.  We propose Uniform Herding, which allocates the current active set across observed classes and uses a bounded candidate pool to refresh their chosen exemplars in the current representation.  On CIFAR-100 with ten class-incremental tasks, a ResNet-18
backbone, active budget $M=2{,}000$, retrieval budget $b=64$, and three seeds,
Uniform Herding obtains $44.00\pm0.51\%$ final average accuracy and
$17.22\pm0.43\%$ forgetting, compared with $42.33\pm1.20\%$ and
  $24.87\pm1.11\%$ for iCaRL.  Within the Uniform Herding protocol, final accuracy decreased when NME or herding was replaced with the tested alternatives, while forgetting increased when distillation was removed. Changing the retrieval budget has a smaller effect across the tested range than changing the active budget. The comparison with iCaRL is end-to-end. It does not isolate the effect of refresh from the other protocol differences. These results are limited to the tested protocol.
\end{abstract}

\input{sections/introduction/introduction.tex}

\input{sections/related_work/related_work.tex}
\input{sections/method/method.tex}
\FloatBarrier
\input{sections/results/results.tex}
\input{sections/discussion/discussion.tex}
\input{sections/conclusion/conclusion.tex}

\clearpage
\bibliographystyle{plainnat}
\bibliography{bib/references}

\clearpage
\appendix
\input{appendix/appendix.tex}

\end{document}

%% file: sections/introduction/introduction.tex
\section{Introduction}
\label{sec:introduction}

In replay-based class-incremental learning, a small exemplar set represents the training history. The backbone changes with each new task, so exemplars selected with an earlier representation may no longer approximate the class distributions well in the current representation. Existing methods combine selection, refresh timing, training objective, and readout into complete protocols; consequently, a performance gap between methods cannot isolate the refresh rule from the other design choices.

We propose Uniform Herding. After each task is completed, it divides the active budget equally among all observed classes and rebuilds each class's exemplar set using greedy herding~\cite{welling2009herding} in the current feature space. Reconstruction candidates come from a bounded candidate pool, while the selected exemplars form the active replay memory.

We then compare Uniform Herding with a faithful implementation of iCaRL~\cite{rebuffi2017icarl} and a
static replay bank on the CIFAR-100 dataset split into ten tasks. Upon arrival, iCaRL herds each new class once and preserves the resulting priority order, retaining the prefix required by the later per-class quota. The comparisons with Uniform Herding are end-to-end: the protocols also differ in training objective, readout, and storage, not only in refresh timing. We also compare Uniform Herding ablations on prediction rule, selection rule, distillation, and classifier head, in addition to active-budget and retrieval sweeps.

At the default budgets ($M{=}2{,}000$ active exemplars, $b{=}64$ retrieval),
Uniform Herding obtains higher mean final accuracy and lower mean forgetting
than both baselines.  Nearest-mean-of-exemplars (NME) prediction and herding selection each outperform
their tested alternatives on mean accuracy; distillation primarily affects
forgetting.  The active-budget sweep changes the metrics more than the
retrieval sweep.

\paragraph{Contributions.}
\begin{itemize}
  \item We define Uniform Herding, which divides the active budget uniformly
  across classes and refreshes the exemplar set from a bounded candidate pool
  in the current representation after each task.
  \item We compare it against iCaRL and a static bank at matched active and
  retrieval budgets, with all protocol differences stated.
  \item We ablate prediction rule, selection rule, distillation, and head
  geometry, and vary active-budget and retrieval sensitivity within the
  proposed protocol.
\end{itemize}

%% file: sections/related_work/related_work.tex
\section{Related Work}
\label{sec:related-work}

\subsection{Class-Incremental Learning}
\label{subsec:related-cil}

Methods for class-incremental learning commonly use parameter regularization, constrained updates, distillation, or replay. EWC and Synaptic Intelligence penalize changes to parameters deemed important for earlier tasks~\cite{kirkpatrick2017ewc,zenke2017si}; GEM and A-GEM constrain gradient updates with episodic memory~\cite{lopezpaz2017gem,chaudhry2019agem}. Without stored data, Learning without Forgetting distills old-task logits from the previous model~\cite{li2016lwf}. Uniform Herding belongs to the replay family and studies how exemplar refresh interacts with the active budget constraint.

\subsection{Replay and Exemplar Selection}
\label{subsec:related-replay}

Experience replay stores a bounded subset of earlier data and mixes it into each
new task's training~\cite{rolnick2019er,chaudhry2019tiny}. iCaRL combines
greedy herding with exemplar replay and nearest-mean-of-exemplars (NME)
prediction~\cite{rebuffi2017icarl}; the herding step greedily builds a set of exemplars whose sample mean tracks the class
mean~\cite{welling2009herding}.  GSS and MIR select exemplars by gradient
diversity and expected interference,
respectively~\cite{aljundi2019gss,aljundi2019mir}.

Uniform Herding follows the same total active budget and greedy herding as
iCaRL but changes the refresh schedule. It utilizes a bounded candidate pool to re-herd every observed class in the current feature space following each task. iCaRL herds a class only once on its arrival, and thereafter truncates the ranked list
as the per-class quota shrinks.  Thus, the comparison is between
two complete protocols, and it does not isolate refresh from the other
differences.

\subsection{Distillation, Classifiers, and Readouts}
\label{subsec:related-readout}
iCaRL distills with sigmoid binary cross-entropy on old-class
targets~\cite{rebuffi2017icarl}; Uniform Herding pairs cross-entropy on all
classes with a temperature-scaled softmax KL term constrained to old classes.

Classifier choice determines the bias between old and new classes.  LUCIR replaces the linear head with a normalized cosine classifier and a margin
loss~\cite{hou2019lucir}; to reduce the new-class bias, BiC and Weight Aligning post-correct the classifier outputs~\cite{wu2019bic,zhao2020wa}.
The learned head is bypassed for prediction by prototype classifiers: iCaRL makes predictions based on exemplar means, and FeCAM expands this using class-covariance weighting~\cite{rebuffi2017icarl,goswami2023fecam}.  Our ablations test
NME vs.\ head-logit prediction, distillation, and head geometry within a
single training protocol.  The evaluation metrics---average accuracy, backward
transfer~\cite{lopezpaz2017gem}, and average
forgetting~\cite{chaudhry2019tiny}---are defined in
Section~\ref{subsec:evaluation-metrics}.

%% file: sections/method/method.tex
\section{Uniform Herding and Experimental Setup}
\label{sec:method}

We evaluated class-incremental learning on the CIFAR-100 dataset with a bounded active exemplar budget. Training runs through tasks $t=0,\ldots,T-1$. Each task brings a disjoint set of $C_{\mathrm{new}}$ classes, and evaluation after task $t$ covers all classes seen so far, without task identity. In the main experiments, $T=10$ and $C_{\mathrm{new}}=10$. We permute the CIFAR-100
classes using split seed 13 and remap them to contiguous internal labels.  From each training class, we keep 30 images for probing and 20 for validation. None enter replay storage.

\input{data/diagram/system_overview.tex}

Figure~\ref{fig:system-overview} summarizes the training, memory-refresh, and evaluation flow.

\subsection{Model and Training}
\label{subsec:model-training}

The learner is a ResNet-18 feature extractor $f_\theta:\mathcal{X}\to
\mathbb{R}^{512}$ with base width 64 and no dropout.  The default classifier
is a cosine-margin classifier. For feature vector $f_\theta(x)$, classifier
weight $w_j$, scale $s$, margin $m$, and target $y$, its logits are
\[
z_j(x;y)=
\begin{cases}
s\left\langle \bar f_\theta(x),\bar w_j\right\rangle-sm, & j=y,\\
s\left\langle \bar f_\theta(x),\bar w_j\right\rangle, & j\neq y,
\end{cases}
\]
where $\bar u=u/\|u\|_2$.  We set $s=30$ and $m=0.35$.  Each new task
increases the classifier head by 10 rows.  Prior to optimization, we imprint the new rows from normalized class-mean features starting with task 1.

We train with SGD at learning rate 0.1, momentum 0.9, weight decay
$5\times10^{-4}$, batch size 128, gradient clipping at 1.0, mixed precision,
and 70 epochs per task.  Training augmentation is random crop (padding 4) and
horizontal flip; evaluation applies normalization only.  Our runs use seeds
1993, 2023, and 42.

\subsection{Active Allocation and Candidate Refresh}
\label{subsec:active-candidate-memory}

Uniform Herding separates the active replay set from the candidate pool used
to refresh it.  Let $\mathcal{C}_t$ be the classes observed after task $t$ and
let $C_t=|\mathcal{C}_t|$.  Given active budget $M$, the quota for the $i$-th
class identifier is
\[
q_{c_i}^{(t)}=
\left\lfloor\frac{M}{C_t}\right\rfloor
+\mathbf{1}\{i<M\bmod C_t\},
\qquad i=0,\ldots,C_t-1.
\]
In our experiments, every class has enough candidates, so the selected sets sum
to $M$ at each completed task boundary.  When a candidate pool falls short of its
quota, we select at most $q_c^{(t)}$ items.

At the start of task $t$, let $\mathcal{B}_c^{(t-1)}$ denote the persistent
candidate pool for an old class and let $\mathcal{P}_{c,\mathrm{cur}}^{(t)}$
denote the raw current-task stream for a newly observed class.  At the task
boundary, the candidate set for class $c$ is
\[
\mathcal{Q}_c^{(t)}=
\mathcal{B}_c^{(t-1)}\cup\mathcal{P}_{c,\mathrm{cur}}^{(t)},
\qquad
\mu_c^{(t)}=\frac{1}{|\mathcal{Q}_c^{(t)}|}
\sum_{x\in\mathcal{Q}_c^{(t)}}f_\theta(x).
\]
For each new class, the persistent pool is empty before its first rebuild. We
retain the current-task stream until the boundary. Once a class has a known
quota, we maintain its persistent candidate pool with class-local bounded reservoir
insertion.

Herding selects the active exemplars from $\mathcal{Q}_c^{(t)}$.  With
$S_0=0$, the $k$-th exemplar is
\[
x_k\in\operatorname*{argmin}_{x\in\mathcal{Q}_c^{(t)}\setminus
\{x_1,\ldots,x_{k-1}\}}
\left\|f_\theta(x)-\left(k\mu_c^{(t)}-S_{k-1}\right)\right\|_2^2,
\qquad S_k=S_{k-1}+f_\theta(x_k).
\]
The selected set is $\mathcal{E}_c^{(t)}=\{x_1,\ldots,x_{q_c^{(t)}}\}$, and
the NME prototype is
\[
\hat\mu_c^{(t)}=\frac{1}{|\mathcal{E}_c^{(t)}|}
\sum_{x\in\mathcal{E}_c^{(t)}}f_\theta(x).
\]
This rebuild runs for every observed class after every task. After selection, the candidate pool keeps the selected items plus a bounded
set of leftover candidates.  At the default
$\rho=\texttt{pool\_multiplier}=3$, each task boundary stores at most $\rho
q_c^{(t)}$ candidates per class and at most $\rho M$ in total in the reported runs.  This bound
excludes the transient full stream of a class before its first rebuild.
$M$ is therefore the active replay budget and $\rho M$ is the boundary
storage candidate bound. The method does not have a total storage budget of $M$.

The candidate pool is a bounded buffer, not a claim of uniform sampling over
the entire historical stream.  Replay draws only from the selected sets.  We use the
default $\rho=3$ and did not sweep it.

\subsection{Replay, Distillation, and Prediction}
\label{subsec:replay-distillation}

At $t=0$, training uses only the current-task minibatches.  For $t\ge1$, each
minibatch is augmented with $b$ exemplars drawn with replacement from
earlier classes.  Sampling is uniform over stored items, not over classes.
Retrieved raw images are re-augmented before concatenation.  We set
$M=2{,}000$ and $b=64$.

The loss combines cross-entropy over the expanded label space with a distillation
on old classes.  We copy a snapshot of the teacher just before expanding the head.
For $t\ge1$,
\[
\begin{aligned}
\mathcal{L}_t
&=\mathcal{L}_{\mathrm{CE}}(z(x;y),y)
+\lambda T_{\mathrm{KD}}^2
\operatorname{KL}\!\left(p^{\mathrm{teach}}(x)\,\middle\|\,p(x)\right),\\
p^{\mathrm{teach}}(x)
&=\operatorname{softmax}\!\left(
z^{\mathrm{teach}}_{0:C_{t-1}-1}(x)/T_{\mathrm{KD}}\right),\\
p(x)
&=\operatorname{softmax}\!\left(
\tilde z_{0:C_{t-1}-1}(x;y)/T_{\mathrm{KD}}\right),
\end{aligned}
\]
where $\tilde z$ is the student's logit vector after undoing the
target-dependent cosine margin for the KD comparison.  We set $\lambda=1$
and $T_{\mathrm{KD}}=2$; at $t=0$ there is no distillation term.

At evaluation, Uniform Herding uses nearest-mean exemplar prediction.  Given
the final representation, the prediction is
\[
\hat y(x)=\operatorname*{argmin}_{c\in\mathcal{C}_{T-1}}
\left\|\bar f_\theta(x)-\bar{\hat\mu}_c^{(T-1)}\right\|_2.
\]
The head-logit rule is used in the corresponding ablation and in the
static-bank baseline's own protocol.  The no-replay baseline has no exemplar
readout.

\subsection{Baselines and Supporting Analyses}
\label{subsec:baselines-supporting}

The iCaRL baseline herds each new class once from its full current-task
pool, stores the prioritized order, and later keeps the prefix required by the
new quota.  It does not re-herd old classes.  It recomputes NME means in
the current feature space over the retained exemplars and discards the
transient current task pool at the task boundary.  Training uses
iCaRL-style sigmoid binary cross-entropy targets for old classes.

The static-bank baseline uses random replacement within class pools, the same
nominal active budget and retrieval count, and cross-entropy training with
head-logit prediction. At task boundaries, it reduces its active store to
the nominal budget, unlike Uniform Herding's additional candidate pool.

Every Uniform Herding ablation changes one choice at a time. We remove KD,
swap NME for head-logit prediction, swap the cosine-margin head for a linear
head, or swap herding for random selection.  Resource sweeps vary
$M\in\{500,2000,4000\}$ and $b\in\{32,64,128\}$ around the default.  T1
compares full protocols.  Among the replay rows, nominal active and retrieval
budgets match, but total retained storage, training objective, and readout
differ. The no-replay baseline has neither replay storage nor retrieval.

\subsection{Evaluation Metrics}
\label{subsec:evaluation-metrics}

After each task $i$, we evaluate separately on every task $j\le i$, producing an
accuracy matrix $R$.  Final average accuracy is
\[
\mathrm{ACC}=\frac{1}{T}\sum_{j=0}^{T-1}R_{T-1,j}.
\]
Forgetting is the mean drop from each old task's best observed accuracy to its
final accuracy:
\[
\mathrm{F}=\frac{1}{T-1}\sum_{j=0}^{T-2}
\left(\max_{i\in\{j,\ldots,T-1\}}R_{i,j}-R_{T-1,j}\right).
\]
Backward transfer is
\[
\mathrm{BWT}=\frac{1}{T-1}\sum_{j=0}^{T-2}
\left(R_{T-1,j}-R_{j,j}\right).
\]
We report all scalar metrics as the mean and population standard deviation
over the three seeds.  For ablation and sensitivity rows, we compute deltas
per seed against the matched Uniform Herding run before averaging.

%% file: data/diagram/system_overview.tex

\definecolor{compfill}{RGB}{240,244,248}
\definecolor{compdraw}{RGB}{140,155,170}
\definecolor{choicefill}{RGB}{255,243,224}
\definecolor{choicedraw}{RGB}{230,126,34}
\definecolor{arrowcol}{RGB}{90,90,90}
\definecolor{notecol}{RGB}{120,120,120}
\definecolor{regionfill}{RGB}{250,250,250}
\definecolor{regiondraw}{RGB}{210,210,210}
\definecolor{labelcol}{RGB}{110,120,130}
\definecolor{mergecol}{RGB}{160,170,180}

\begin{figure}[tbp]
\centering

\begin{tikzpicture}[
  scale=0.9,
  >=Stealth,
  base/.style={
    draw, rounded corners=2pt, minimum height=7.5mm,
    minimum width=24mm, align=center, inner sep=2.5pt,
    font=\sffamily\scriptsize, line width=0.5pt
  },
  comp/.style={base, fill=compfill, draw=compdraw},
  choice/.style={base, fill=choicefill, draw=choicedraw, line width=0.7pt},
  flow/.style={->, line width=0.55pt, draw=arrowcol},
  back/.style={flow, dashed},
  note/.style={font=\sffamily\tiny, text=notecol, align=center, text width=23mm},
  region/.style={
    draw=regiondraw, fill=regionfill, rounded corners=4pt,
    inner sep=8pt, line width=0.3pt
  },
  rlabel/.style={font=\sffamily\scriptsize\bfseries, text=labelcol},
]

\node[comp] (data) at (0, 0)       {Task $t$ data};
\node[comp] (rpl)  at (0, -1.05)   {Replay from active set};
\node[circle, fill=mergecol, inner sep=1.5pt]
             (mrg) at (2.8, -0.52) {};
\node[comp] (bb)   at (5.2, -0.52)   {Backbone $f_\theta$};
\node[choice] (head) at (8.2, -0.52) {Head};
\node[choice] (loss) at (11.2, -0.52) {Loss};
\node[comp] (tch)  at (11.2, -1.55) {Teacher logits};

\draw[flow] (data.east) -- (mrg.west);
\draw[flow] (rpl.east)  -- (mrg.west);
\draw[flow] (mrg.east)  -- (bb.west);
\draw[flow] (bb.east)   -- (head.west);
\draw[flow] (head.east) -- (loss.west);
\draw[flow] (tch.north) -- (loss.south);

\node[note, anchor=north west] (rplnote) at ($(rpl.south) + (-10pt, -2pt)$) {$b$ samples per minibatch};
\node[note, below=2pt of head] (headnote) {cosine-margin / linear};
\node[note, above=2pt of loss] (lossnote) {CE + KD / CE only};

\coordinate (R1pad) at ($(data.north)+(0,0.42)$);

\node[comp] (cand) at (0, -3.45) {Candidate pools $\mathcal{Q}_c$};
\node[choice] (sel) at (3.35, -3.45) {Herding / random};
\node[comp] (active) at (6.75, -3.45) {Active set $\mathcal{E}$};
\node[comp] (proto) at (10.7, -3.45) {NME prototypes};

\draw[flow] (cand.east) -- (sel.west);
\draw[flow] (sel.east) -- (active.west);
\draw[flow] (active.east) -- (proto.west);

\node[note, below=2pt of cand] {current stream + persistent candidates};
\node[note, below=2pt of sel] {rebuild in current $f_\theta$};
\node[note, below=2pt of active] {active budget $M$; replay source};
\node[note, below=2pt of proto, text width=17mm] {selected-exemplar means};

\coordinate (R2padT) at ($(cand.north)+(0,0.35)$);
\coordinate (R2padB) at ($(active.south)+(0,-0.68)$);

\node[comp] (qry)  at (0, -5.95)    {Query $x$};
\node[comp] (bb2)  at (3.35, -5.95) {Backbone $f_\theta$};
\node[choice] (rdo) at (6.75, -5.95) {Readout};
\node[comp] (pred) at (10.7, -5.95)  {$\hat{y}$};

\draw[flow] (qry.east) -- (bb2.west);
\draw[flow] (bb2.east) -- (rdo.west);
\draw[flow] (rdo.east) -- (pred.west);

\node[note, below=2pt of rdo] {NME / head-logit};

\coordinate (R3padT) at ($(qry.north)+(0,0.42)$);
\coordinate (R3padB) at ($(rdo.south)+(0,-0.42)$);

\draw[back] (active.north) -- ++(0, 0.5) -| (rpl.south);
\draw[back] ($(proto.south)+(-1.15,0)$) -- ++(0, -1.15) -| (rdo.north);

\begin{scope}[on background layer]
  \node[region, fit=(data)(rpl)(loss)(tch)(R1pad)(rplnote)(headnote)(lossnote)] (R1) {};
  \node[region, fit=(cand)(proto)(R2padT)(R2padB)] (R2) {};
  \node[region, fit=(qry)(pred)(R3padT)(R3padB)] (R3) {};
\end{scope}

\node[rlabel, anchor=north west]
  at ([shift={(5pt,-5pt)}]R1.north west)
  {Training ($t\geq1$)};
\node[rlabel, anchor=north west]
  at ([shift={(5pt,-5pt)}]R2.north west)
  {Memory refresh (after each task)};
\node[rlabel, anchor=north west]
  at ([shift={(5pt,-5pt)}]R3.north west)
  {Evaluation};

\end{tikzpicture}

\caption{Overview of Uniform Herding. At each task boundary, class-specific
candidate pools are refreshed in the current representation and herding selects
the active replay set with budget $M$. Replay draws only from this active set,
while candidate storage is bounded by $\rho M$ at completed boundaries; the
retrieval count is $b$ per minibatch. Orange boxes indicate choices examined in
the within-method comparisons. Dashed arrows show the replay and prototype
feedback paths.}
\label{fig:system-overview}
\end{figure}
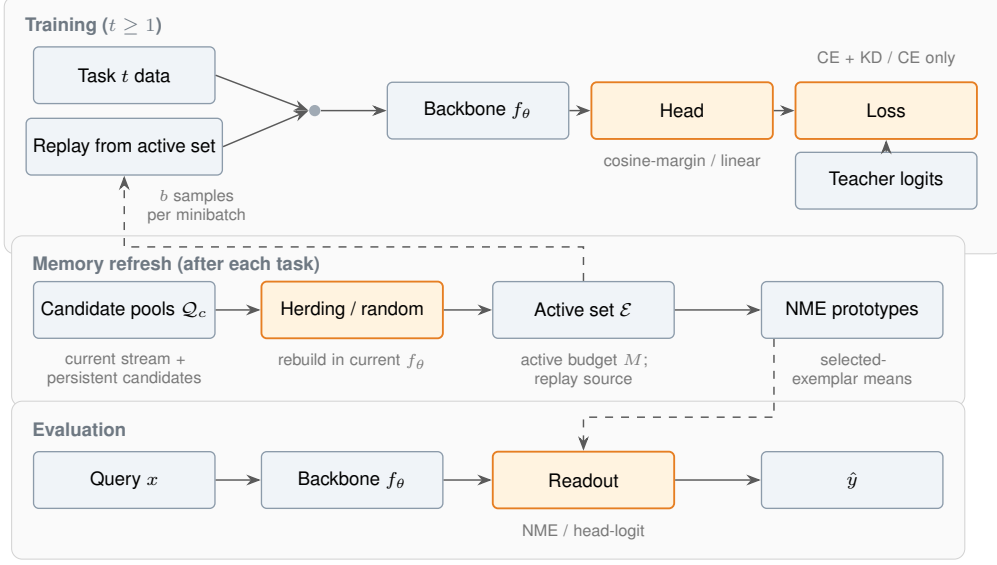

%% file: sections/results/results.tex
\section{Results}
\label{sec:results}

All results follow the protocol in Section~\ref{sec:method} and are reported
as mean $\pm$ population standard deviation over three seeds.  The main
comparison in Table~\ref{tab:t1-master-results} uses active budget
$M=2{,}000$ and retrieval budget $b=64$.

\subsection{Main Method Comparison}
\label{subsec:main-comparison}

\begin{figure}[htbp]
  \centering
  \includegraphics[width=0.92\linewidth]{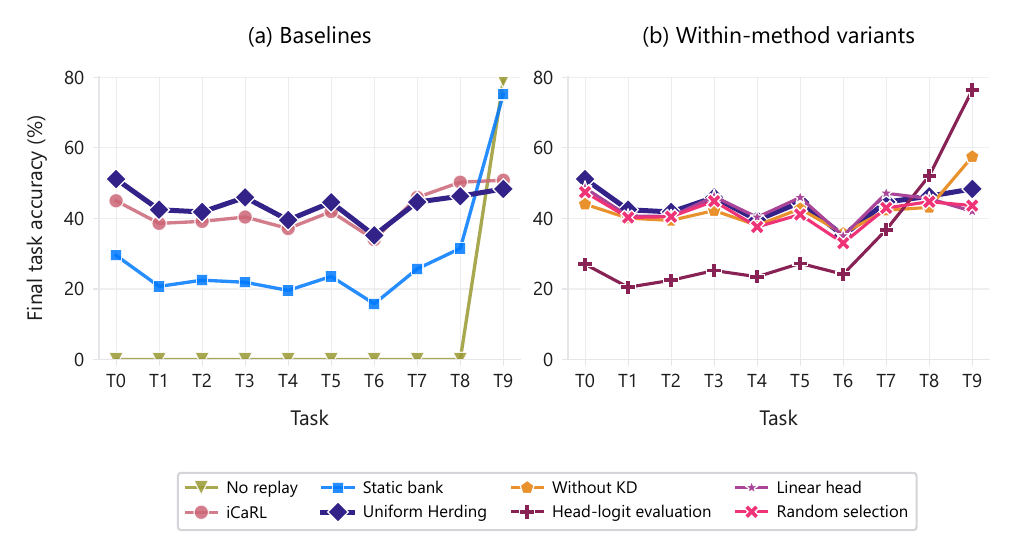}
  \caption{Final per-task accuracies. Panel (a) compares the baselines with
  Uniform Herding. Panel (b) shows the within-method variants.}
  \label{fig:per-task-accuracy}
\end{figure}

At the common nominal active budget, Uniform Herding obtains
$44.00\pm0.51\%$ final average accuracy and $17.22\pm0.43\%$ forgetting.
iCaRL reaches $42.33\pm1.20\%$ accuracy and $24.87\pm1.11\%$ forgetting. 
The static bank achieves $28.60\pm1.35\%$ and $55.86\pm1.42\%$.
The comparison is end-to-end. The baselines differ in candidate retention and
training objective, and the static bank also uses a different readout. The
appendix details
these protocol differences.  Figure~\ref{fig:per-task-accuracy}
shows final per-task accuracies.

\begin{table}[htbp]
  \centering
  \caption{Main comparison on CIFAR-100. All methods use a nominal active
  exemplar budget of 2,000 and retrieval budget of 64. Uniform Herding also
  retains candidate storage for refresh.}
  \label{tab:t1-master-results}
  \small
  \resizebox{\linewidth}{!}{%
  \begin{tabular}{lrrr}
    \toprule
    Method & Average accuracy (\%) & Forgetting (\%) & BWT (\%) \\
    \midrule
    iCaRL & $42.33\pm1.20$ & $24.87\pm1.11$ & $-24.87\pm1.11$ \\
    Static bank & $28.60\pm1.35$ & $55.86\pm1.42$ & $-55.86\pm1.42$ \\
    Uniform Herding & $44.00\pm0.51$ & $17.22\pm0.43$ & $-16.91\pm0.41$ \\
    \bottomrule
  \end{tabular}
  }
\end{table}

\subsection{Supporting Analyses of Uniform Herding}
\label{subsec:supporting-analyses}

Table~\ref{tab:t2-component-ablations} reports matched within-method ablations
of the prediction rule, selection rule, distillation, and classifier head.
Replacing NME with head-logit evaluation drops accuracy by $10.46$~pp and
raises forgetting by $30.36$~pp.  Replacing
herding with random selection costs $2.39$~pp in accuracy and $1.16$~pp in
forgetting.  Removing KD decreases accuracy by $1.48$~pp but increases
forgetting by $11.66$~pp. Replacing the cosine-margin head with a linear head
decreases accuracy by $0.79$~pp and increases forgetting by $0.12$~pp.  These
comparisons describe the current configuration; they do not establish effects beyond it.
Appendix~Figure~\ref{fig:app-component-attribution} shows the per-seed deltas.

\begin{figure}[htbp]
  \centering
  \includegraphics[width=\linewidth]{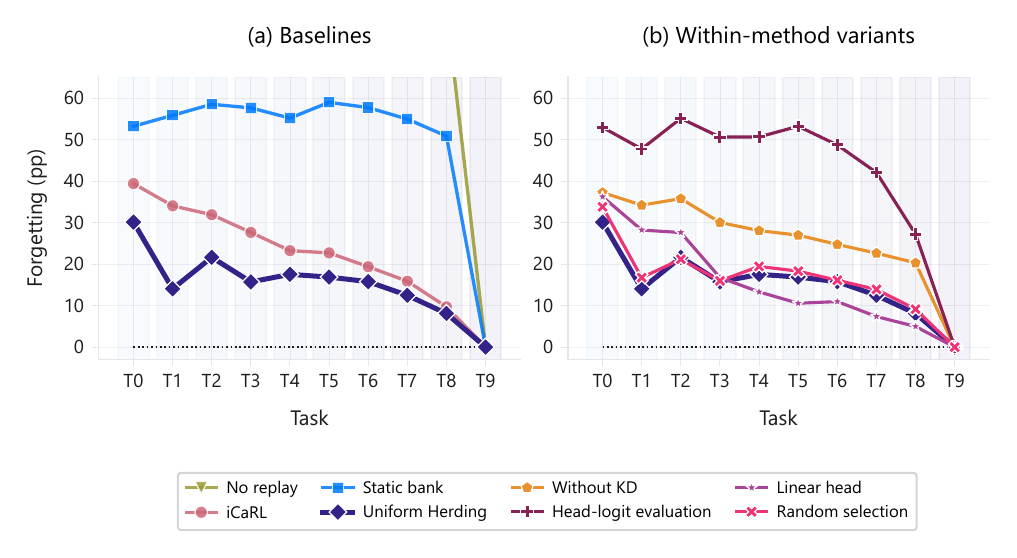}
  \caption{Forgetting by task index for Uniform Herding and two adverse
  configurations. The final task has zero forgetting by definition because no
  subsequent task is learned.}
  \label{fig:forgetting-by-age}
\end{figure}

Figure~\ref{fig:forgetting-by-age} plots forgetting by task. From tasks 0--8,
forgetting is $51$--$59$~pp for the static bank and $27$--$55$~pp for head-logit
evaluation, compared to $8.1$--$30.1$~pp for Uniform Herding. The pattern is consistent with recency bias: both unfavorable settings maintain good accuracy on the most recent task despite their aggregate forgetting. Task 9 has zero forgetting by definition since
no subsequent training degrades it.

\begin{table}[htbp]
  \centering
  \caption{Matched within-method comparisons for Uniform Herding. Deltas are
  paired by seed against the main configuration.}
  \label{tab:t2-component-ablations}
  \small
  \resizebox{\linewidth}{!}{%
  \begin{tabular}{lrrrr}
    \toprule
    Variant & Average accuracy (\%) & Forgetting (\%) & $\Delta$ accuracy (pp) & $\Delta$ forgetting (pp) \\
    \midrule
    Without KD & $42.52\pm0.61$ & $28.88\pm0.19$ & $-1.48$ & $+11.66$ \\
    Head-logit evaluation & $33.54\pm0.39$ & $47.58\pm0.84$ & $-10.46$ & $+30.36$ \\
    Linear head & $43.22\pm0.36$ & $17.34\pm0.68$ & $-0.79$ & $+0.12$ \\
    Random selection & $41.61\pm0.53$ & $18.38\pm0.47$ & $-2.39$ & $+1.16$ \\
    \bottomrule
  \end{tabular}
  }
\end{table}

\subsection{Active-Budget and Retrieval Sensitivity}
\label{subsec:resource-sensitivity}

Because Uniform Herding uses $\rho=3$, changing the active budget
changes the candidate-pool capacity as well.  Reducing $M$ from 2,000 to 500
lowers accuracy by $10.05$~pp and raises forgetting by $13.26$~pp.  Increasing
$M$ to 4,000 raises accuracy by $3.32$~pp and lowers forgetting by
$3.91$~pp.  Three tested points do not establish a scaling law.

\begin{figure}[htbp]
  \centering
  \includegraphics[width=0.92\linewidth]{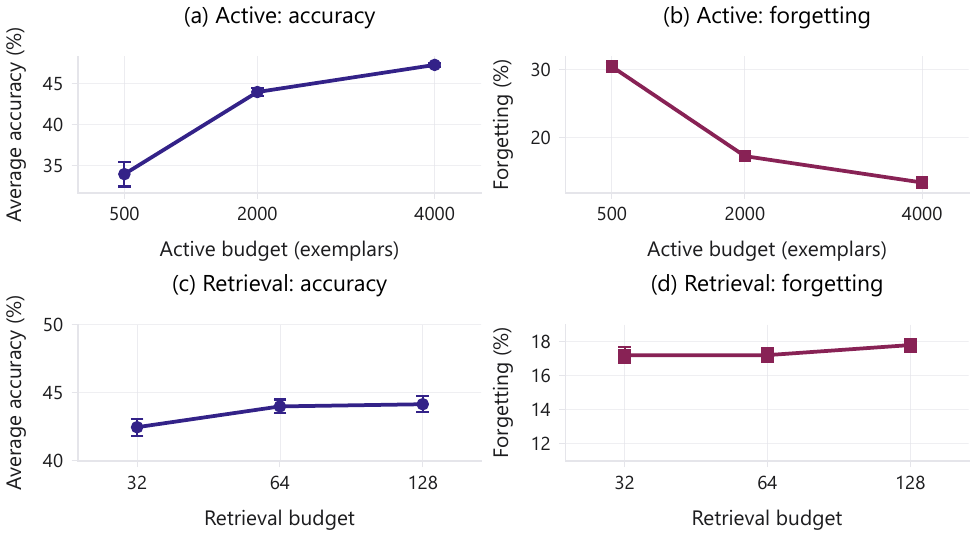}
  \caption{Sensitivity to active exemplar budget and replay retrieval budget.
  Error bars denote population standard deviation over three seeds.}
  \label{fig:resource-sensitivity}
\end{figure}

With $M=2{,}000$ fixed, reducing retrieval from 64 to 32 decreases accuracy by
$1.53$~pp and decreases forgetting by $0.01$~pp.  Increasing retrieval to 128 increases
accuracy by $0.16$~pp and increases forgetting by $0.59$~pp.  The retrieval sweep
produces a smaller mean change than the active-budget sweep
(Table~\ref{tab:t3-resource-sensitivity}).
Figure~\ref{fig:resource-sensitivity} plots both sweeps.

\begin{table}[htbp]
  \centering
  \caption{Resource sensitivity relative to Uniform Herding at $M=2{,}000$ and
  retrieval budget 64. Active-budget changes also change the candidate-pool
  capacity through $\rho=3$; deltas are paired by seed.}
  \label{tab:t3-resource-sensitivity}
  \small
  \resizebox{\linewidth}{!}{%
  \begin{tabular}{llrrrr}
    \toprule
    Axis & Value & Average accuracy (\%) & Forgetting (\%) & $\Delta$ accuracy (pp) & $\Delta$ forgetting (pp) \\
    \midrule
    Active budget & 500 & $33.95\pm1.51$ & $30.48\pm0.54$ & $-10.05$ & $+13.26$ \\
    Active budget & 2,000 & $44.00\pm0.51$ & $17.22\pm0.43$ & $0.00$ & $0.00$ \\
    Active budget & 4,000 & $47.32\pm0.24$ & $13.31\pm0.59$ & $+3.32$ & $-3.91$ \\
    Retrieval & 32 & $42.47\pm0.63$ & $17.22\pm0.47$ & $-1.53$ & $-0.01$ \\
    Retrieval & 64 & $44.00\pm0.51$ & $17.22\pm0.43$ & $0.00$ & $0.00$ \\
    Retrieval & 128 & $44.16\pm0.59$ & $17.81\pm0.38$ & $+0.16$ & $+0.59$ \\
    \bottomrule
  \end{tabular}
  }
\end{table}

%% file: sections/discussion/discussion.tex
\section{Discussion}
\label{sec:discussion}

Uniform Herding achieves higher mean final accuracy and lower mean forgetting
than the faithful iCaRL implementation and the static-bank baseline in the evaluated protocol.
This is a complete-method result. Because the protocols differ in refresh policy,
training objective, and retained candidate storage and the static bank also uses a different
readout, the comparison does not estimate the effect of refresh alone.

\subsection{Evidence About the Proposed Configuration}
\label{subsec:what-drives-performance}

Within Uniform Herding, NME evaluation is strongly favored over head-logit evaluation
by $10.46$~pp in accuracy and $30.36$~pp in forgetting.  Since both
variants share the training procedure, the gap points to the value of
matching the readout to the selected exemplar representation in this protocol.

Herding selection yields $2.39$~pp higher accuracy and reduces mean forgetting
by $1.16$~pp relative to random selection; this small difference is not estimated
reliably with three seeds. Removing KD decreases accuracy by
$1.48$~pp but increases forgetting by $11.66$~pp, suggesting that
distillation serves a retention role in this objective.  Replacing the
cosine-margin head with a linear head produces the smallest shift among
these ablations ($-0.79$~pp accuracy, $+0.12$~pp forgetting).  These
comparisons describe the current configuration and they do not establish effects beyond it.

\subsection{Resource and Storage Implications}
\label{subsec:resource-allocation}

Reducing the active budget from 2,000 to 500 lowers accuracy by $10.05$~pp and
raises forgetting by $13.26$~pp.  Raising retrieval from 64 to 128 increases
mean accuracy by $0.16$~pp but also increases mean forgetting by $0.59$~pp.
Because $\rho=3$, the active-budget sweep simultaneously changes candidate
capacity. Uniform Herding therefore trades additional candidate storage and
refresh computation for an active replay set of size $M$. The comparison with
iCaRL or the static bank is not storage-matched. The observed pattern
is conditional on the backbone, data stream, and budget range.

\subsection{Limitations}
\label{subsec:limitations}

All experiments use the CIFAR-100 dataset, one ten-task partition (split seed 13), one
ResNet-18 backbone, and three training seeds.  No variation in class order,
task granularity, dataset, or architecture is tested.  The active-budget and
retrieval sweeps cover only the reported values, and the default $\rho=3$ was
not swept separately.  A new class retains its full transient stream before
its first rebuild, so the boundary candidate storage bound does not capture all
mid-task storage.

The iCaRL comparison does not isolate refresh from the objective and storage
differences between the two protocols.  No matched experiment varies only the refresh policy while retaining the objective, head, readout, active budget, candidate multiplier, data order, and seed set fixed. Three matched seeds are used in the paired supporting comparisons, so tiny effects are not estimated reliably.

The next experiment should use one common objective, head, readout, active
budget, candidate multiplier, data order, and seed set, then replace
arrival-time prefix truncation with refresh-all-class candidate reselection.
That design would support a refresh-specific claim.

%% file: sections/conclusion/conclusion.tex
\section{Conclusion}
\label{sec:conclusion}

We proposed Uniform Herding, a replay method that allocates a uniform active
exemplar budget and refreshes every observed class in the current feature representation.
Selected exemplars form the replay set, while a bounded candidate pool retains
examples for future reselection. At $M=2{,}000$ and $b=64$, Uniform Herding
reaches $44.00\pm0.51\%$ final average accuracy and $17.22\pm0.43\%$
forgetting, against $42.33\pm1.20\%$ and $24.87\pm1.11\%$ for iCaRL,
and $28.60\pm1.35\%$ and $55.86\pm1.42\%$ for the static bank.

Within the proposed configuration, NME and herding selection yield higher mean final
accuracy than the alternatives evaluated, and distillation mainly improves
retention.  The active-budget sweep shifts both metrics more than the retrieval
sweep over the reported values.  These findings are conditional on
the evaluated protocol and do not show that refresh alone causes the iCaRL gap.

%% file: appendix/appendix.tex
\section{Reproducibility Details}
\label{app:reproducibility}

The source code to reproduce all experiments is available at \url{https://github.com/neryva-lab/uniform-herding}.

All experiments used CIFAR-100 in a fixed ten-task class-incremental
partition of ten classes per task.  The class partition was created by split
seed 13 and shared by all runs. The three reported seeds therefore vary
training randomness rather than class order.  Each task used a ResNet-18 with
64 base filters and no dropout, batch size 128, mixed-precision arithmetic, and
SGD with learning rate 0.1, momentum 0.9, weight decay $5\times10^{-4}$, and
gradient clipping at 1.0.  No learning-rate schedule or warm-up was used.  The
configured task length was 70 epochs.  Run metadata records 71 epochs for every
seed because the trainer's epoch counter was incremented once after the final
configured epoch. This recording discrepancy does not alter the configured
schedule.

Until a variant changes a setting, Uniform Herding uses a
cosine-margin head (scale 30, margin 0.35), weight imprinting for each new
task's head rows after task 0, distillation weight 1 at temperature 2, and
retrieval budget 64.  The no-distillation ablation sets the distillation weight
to zero. The linear-head variant replaces the cosine-margin head.  The static
bank and head-logit variant use head-logit prediction. All replay
configurations other than these use NME prediction.  The no-replay baseline has
no replay readout.  Uniform Herding replays only its selected exemplars.  Its
implementation default is $\rho=\texttt{pool\_multiplier}=3$:
the candidate pool is bounded by $\rho M$ at completed task boundaries in the
reported setting, while the current-task stream before a class's first rebuild
is transient and may exceed that bound.

\begin{table}[H]
  \centering
  \caption{Protocol settings shared across runs unless an ablation explicitly
  changes them.}
  \label{tab:app-protocol}
  \small
  \resizebox{\linewidth}{!}{%
  \begin{tabular}{ll}
    \toprule
    Setting & Value \\
    \midrule
    Dataset and partition & CIFAR-100; 10 tasks $\times$ 10 classes; split seed 13 \\
    Probe / validation split & 30 / 20 \\
    Backbone & ResNet-18; 64 base filters; dropout 0 \\
    Optimizer & SGD; lr 0.1; momentum 0.9; weight decay $5\times10^{-4}$ \\
    Training & 70 configured epochs per task; batch size 128; gradient clip 1.0 \\
    Precision & 16-mixed \\
    Random seeds & 1993, 2023, 42 \\
    Active exemplar budget & 2,000 for replay configurations; not used by no replay \\
    Retrieval budget & 64 for replay configurations; not used by no replay \\
    Candidate multiplier ($\rho$) & 3 for Uniform Herding; implementation default \\
    Replay source & Selected active exemplars; candidate pool used only for refresh \\
    Hardware & Tesla T4 \\
    Software & Python 3.12.13; PyTorch 2.11.0+cu128; Lightning 2.6.5 \\
    \bottomrule
  \end{tabular}
  }
\end{table}

\section{Task-Level Results}
\label{app:task-results}

\begin{figure}[htbp]
  \centering
  \includegraphics[width=\linewidth]{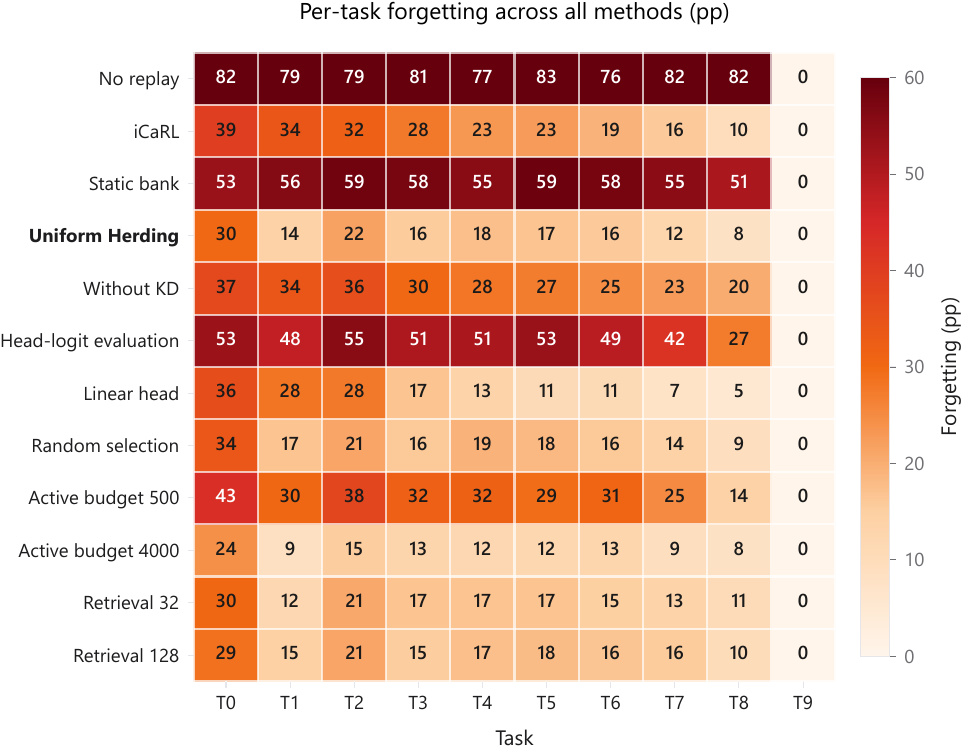}
  \caption{Per-task forgetting, in percentage points, for all evaluated
  configurations. Each entry is the drop from the task's best observed accuracy
  to its final accuracy. Task T9 is zero by definition because it is introduced
  last.}
  \label{fig:app-forgetting}
\end{figure}

For task $i$, reported forgetting is the difference between the best accuracy
attained on task $i$ during training and its accuracy after the final task.
Figure~\ref{fig:app-forgetting} gives this quantity for every evaluated
configuration and task.
The final task has zero forgetting by construction, so the figure separates
aggregate retention from its distribution over task.

\begin{figure}[htbp]
  \centering
  \begin{minipage}[t]{0.48\linewidth}
    \centering
    \includegraphics[width=\linewidth]{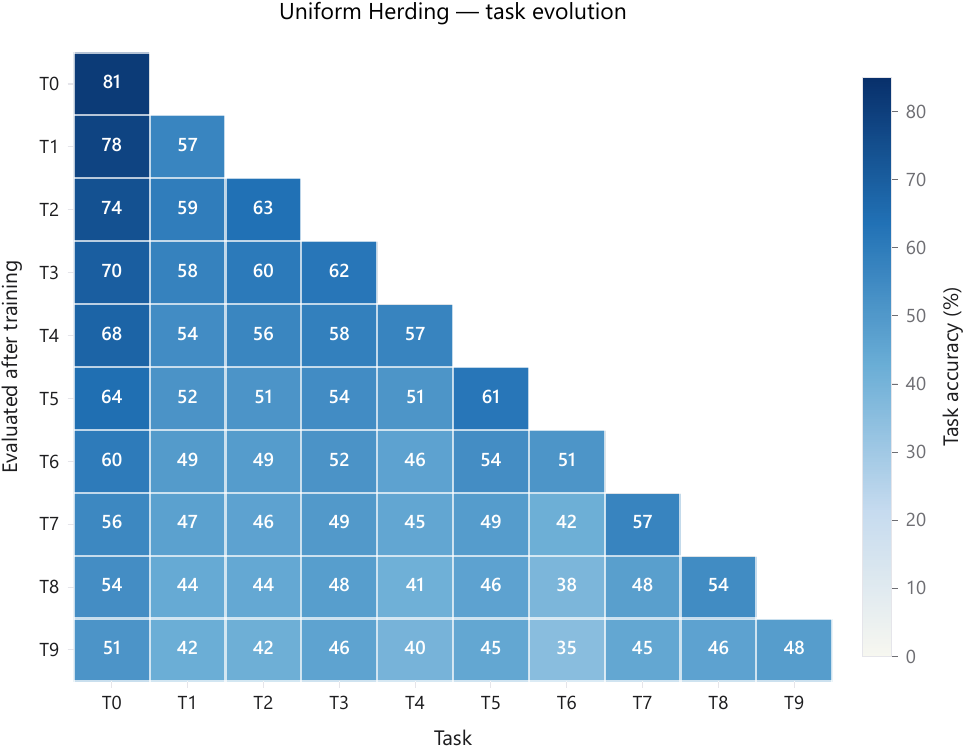}
    \caption{Task-evolution accuracy matrix for Uniform Herding with NME prediction.}
    \label{fig:app-uniform-evolution}
  \end{minipage}\hfill
  \begin{minipage}[t]{0.48\linewidth}
    \centering
    \includegraphics[width=\linewidth]{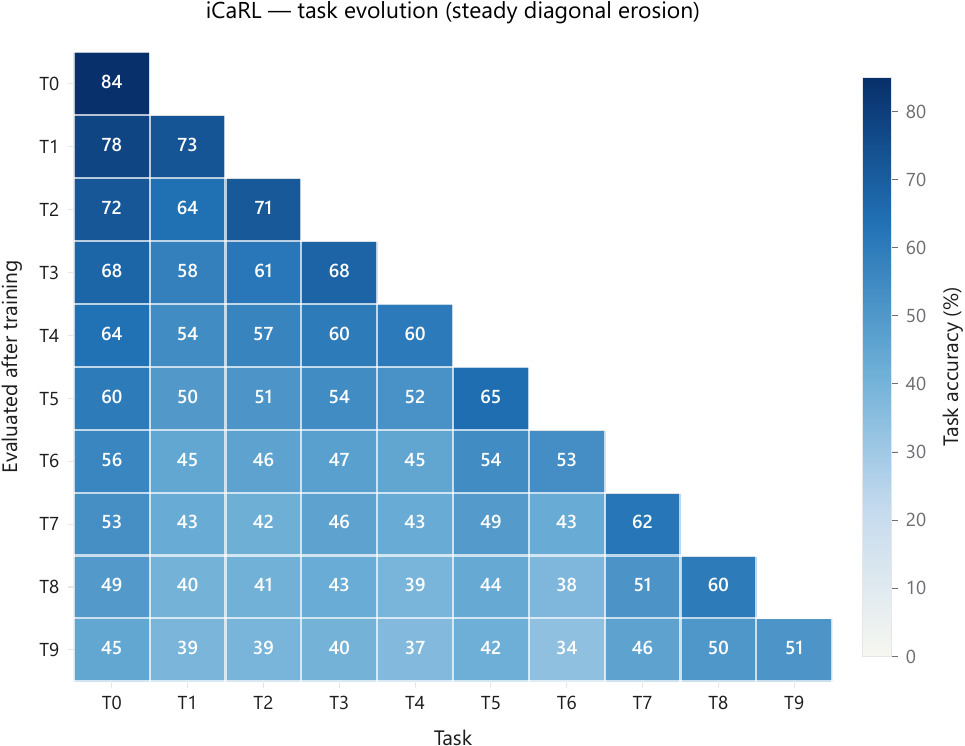}
    \caption{Task-evolution accuracy matrix for iCaRL.}
    \label{fig:app-icarl-evolution}
  \end{minipage}
\end{figure}

Figures~\ref{fig:app-uniform-evolution} and~\ref{fig:app-icarl-evolution}
show the complete task-by-time accuracy matrices for Uniform Herding and
iCaRL.  Row $i$, column $t$ is the accuracy on task $i$ after learning through
task $t$; cells with $i>t$ are undefined and omitted.  These matrices are the
underlying trajectories from which task-level forgetting is calculated.

\begin{figure}[htbp]
  \centering
  \includegraphics[width=\linewidth]{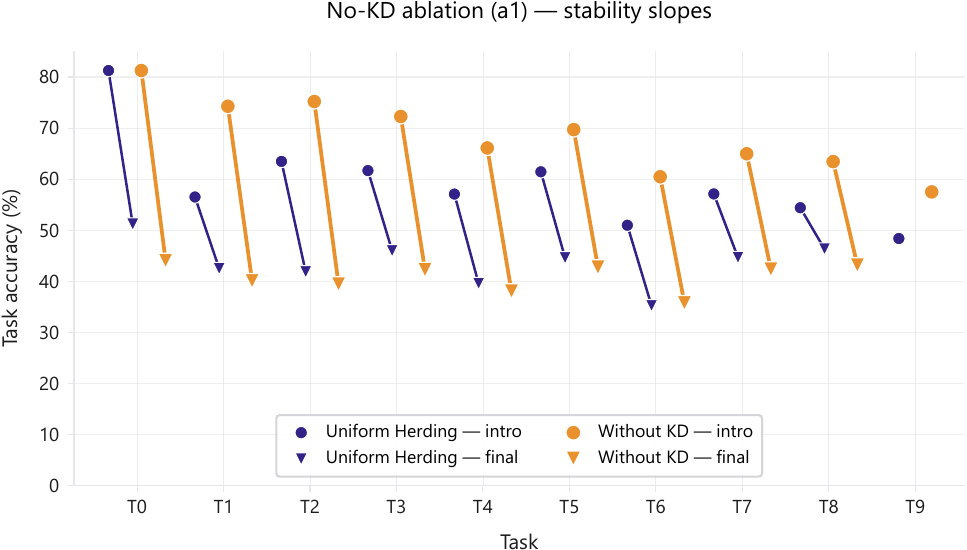}
  \caption{Seed-level task trajectories for Uniform Herding and the
  no-distillation ablation. Lines connect task-introduction accuracy to final
  accuracy for the same task and seed.}
  \label{fig:app-kd-slopes}
\end{figure}

Figure~\ref{fig:app-kd-slopes} provides a seed-level view of the stability
comparison.  For each earlier task and seed, it connects accuracy at
introduction with final accuracy.  It visualizes the per-seed quantities used
for the stability comparison and is not an additional statistical test.

\section{Additional Diagnostics}
\label{app:additional-diagnostics}

\begin{figure}[htbp]
  \centering
  \includegraphics[width=\linewidth]{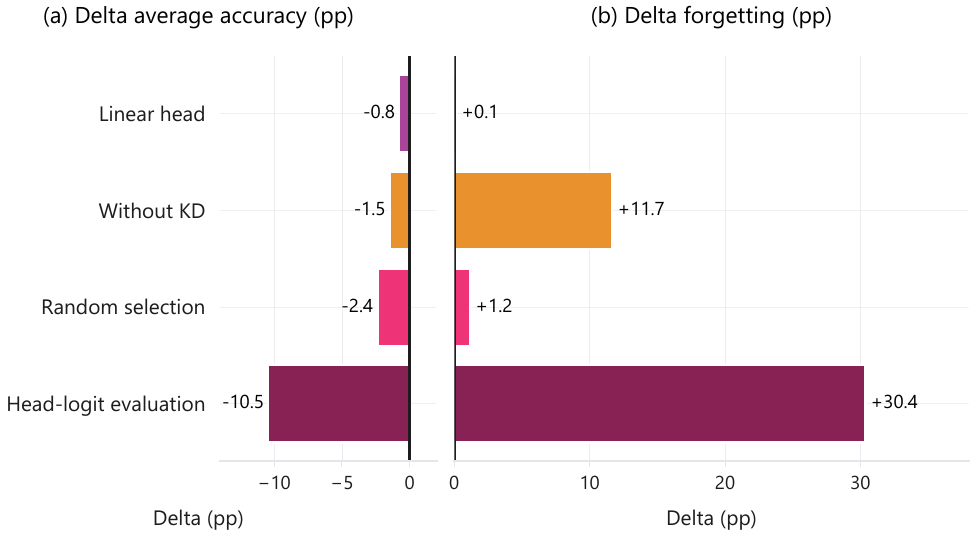}
  \caption{Matched per-seed changes relative to Uniform Herding. Negative
  accuracy changes and positive forgetting changes are unfavorable.}
  \label{fig:app-component-attribution}
\end{figure}

\begin{figure}[htbp]
  \centering
  \includegraphics[width=\linewidth]{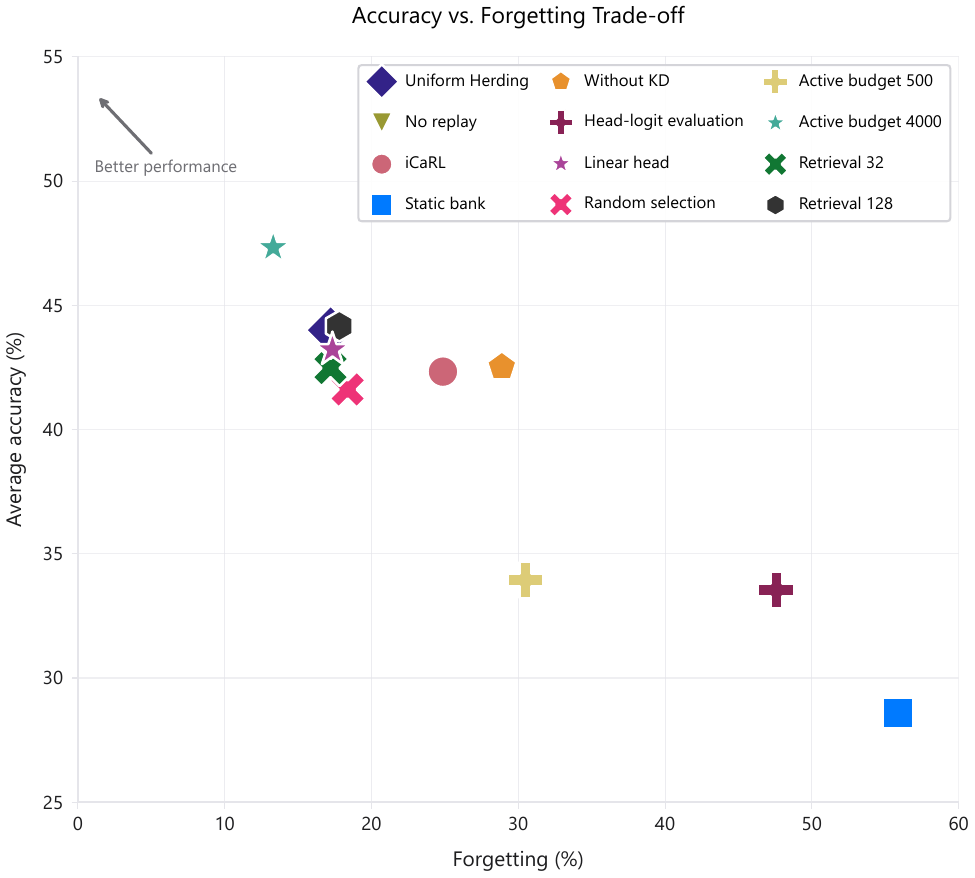}
  \caption{Average accuracy versus forgetting for baselines, within-method
  variants, and resource configurations. Higher accuracy and lower forgetting
  are preferred. The points do not define a causal trade-off curve.}
  \label{fig:app-accuracy-forgetting}
\end{figure}

The following figures provide supplementary views of the matched within-method
comparisons (Figure~\ref{fig:app-component-attribution}) and the joint
accuracy--forgetting distribution
(Figure~\ref{fig:app-accuracy-forgetting}).  The exact
per-configuration values remain in Tables~\ref{tab:t2-component-ablations}
and~\ref{tab:t3-resource-sensitivity}.

\clearpage

\section{Exemplar Quotas}
\label{app:quotas}

After task $t$, the active budget is divided among the $10(t+1)$ observed
classes.  Table~\ref{tab:app-quotas} lists the resulting per-class quotas.
When the budget is not divisible by the number of observed classes, floor
rounding assigns the remainder to the first classes in identifier order.

\begin{table}[htbp]
  \centering
  \caption{Per-class active exemplar quota after each task. Ranges indicate
  the one-exemplar difference induced by floor rounding.}
  \label{tab:app-quotas}
  \small
  \resizebox{\linewidth}{!}{%
  \begin{tabular}{lrrrrrrrrrr}
    \toprule
    Active budget & T0 & T1 & T2 & T3 & T4 & T5 & T6 & T7 & T8 & T9 \\
    \midrule
    500 & 50 & 25 & 16--17 & 12--13 & 10 & 8--9 & 7--8 & 6--7 & 5--6 & 5 \\
    2,000 & 200 & 100 & 66--67 & 50 & 40 & 33--34 & 28--29 & 25 & 22--23 & 20 \\
    4,000 & 400 & 200 & 133--134 & 100 & 80 & 66--67 & 57--58 & 50 & 44--45 & 40 \\
    \bottomrule
  \end{tabular}
  }
\end{table}

\section{Seed-Level Outcomes and Compute}
\label{app:seed-compute}

Table~\ref{tab:app-seeds} reports the raw aggregate metrics for every run and
seed.  Values are in percentages.  The table makes the reported means and
standard deviations auditable without treating the three training seeds as
independent task-level observations.

\begin{table}[htbp]
  \centering
  \caption{Per-seed final average accuracy, forgetting, and backward transfer.}
  \label{tab:app-seeds}
  \scriptsize
  \begin{tabular}{llrrr}
    \toprule
    ID & Configuration / seed & Average accuracy & Forgetting & BWT \\
    \midrule
    B0 & No replay / 1993 & 7.79 & 80.73 & -80.73 \\
    B0 & No replay / 2023 & 8.05 & 78.94 & -78.94 \\
    B0 & No replay / 42 & 7.89 & 80.97 & -80.97 \\
    B1 & iCaRL / 1993 & 40.88 & 25.83 & -25.83 \\
    B1 & iCaRL / 2023 & 43.82 & 25.46 & -25.46 \\
    B1 & iCaRL / 42 & 42.29 & 23.32 & -23.32 \\
    B2 & Static bank / 1993 & 28.15 & 56.34 & -56.34 \\
    B2 & Static bank / 2023 & 30.43 & 53.93 & -53.93 \\
    B2 & Static bank / 42 & 27.22 & 57.31 & -57.31 \\
    B3 & Uniform Herding / 1993 & 43.64 & 16.68 & -16.41 \\
    B3 & Uniform Herding / 2023 & 44.72 & 17.26 & -16.91 \\
    B3 & Uniform Herding / 42 & 43.65 & 17.73 & -17.41 \\
    a1 & No KD / 1993 & 41.72 & 29.10 & -29.10 \\
    a1 & No KD / 2023 & 42.66 & 28.90 & -28.90 \\
    a1 & No KD / 42 & 43.19 & 28.64 & -28.64 \\
    a2 & Head-logit / 1993 & 34.09 & 46.39 & -46.39 \\
    a2 & Head-logit / 2023 & 33.34 & 48.12 & -48.12 \\
    a2 & Head-logit / 42 & 33.20 & 48.22 & -48.22 \\
    a3 & Linear head / 1993 & 43.32 & 17.73 & -17.73 \\
    a3 & Linear head / 2023 & 43.60 & 17.91 & -17.87 \\
    a3 & Linear head / 42 & 42.73 & 16.39 & -16.39 \\
    a4 & Random selection / 1993 & 40.87 & 18.59 & -18.59 \\
    a4 & Random selection / 2023 & 42.03 & 18.82 & -18.82 \\
    a4 & Random selection / 42 & 41.94 & 17.73 & -17.43 \\
    s1 & Active budget 500 / 1993 & 32.08 & 31.18 & -31.18 \\
    s1 & Active budget 500 / 2023 & 35.79 & 29.86 & -29.86 \\
    s1 & Active budget 500 / 42 & 33.98 & 30.41 & -30.41 \\
    s2 & Active budget 4000 / 1993 & 47.03 & 12.52 & -12.02 \\
    s2 & Active budget 4000 / 2023 & 47.61 & 13.93 & -13.59 \\
    s2 & Active budget 4000 / 42 & 47.33 & 13.48 & -12.62 \\
    s3 & Retrieval 32 / 1993 & 41.93 & 16.91 & -16.79 \\
    s3 & Retrieval 32 / 2023 & 43.36 & 17.88 & -17.41 \\
    s3 & Retrieval 32 / 42 & 42.13 & 16.86 & -16.40 \\
    s4 & Retrieval 128 / 1993 & 43.41 & 17.48 & -17.21 \\
    s4 & Retrieval 128 / 2023 & 44.84 & 18.34 & -18.01 \\
    s4 & Retrieval 128 / 42 & 44.24 & 17.62 & -17.23 \\
    \bottomrule
  \end{tabular}
\end{table}

Table~\ref{tab:app-compute} reports elapsed wall-clock time per run on a Tesla
T4.  These timings describe the implementation and hardware used here. They
are not hardware-independent efficiency comparisons.

\begin{table}[htbp]
  \centering
  \caption{Wall-clock time per run in seconds (mean $\pm$ standard deviation
  across three seeds), measured on a Tesla T4.}
  \label{tab:app-compute}
  \small
  \begin{tabular}{llr}
    \toprule
    ID & Configuration & Seconds \\
    \midrule
    B0 & No replay & $2022.4 \pm 79.4$ \\
    B1 & iCaRL & $3649.3 \pm 26.1$ \\
    B2 & Static bank & $2756.4 \pm 109.3$ \\
    B3 & Uniform Herding & $4635.4 \pm 47.5$ \\
    a1 & No KD & $3446.0 \pm 43.7$ \\
    a2 & Head-logit & $4332.4 \pm 33.3$ \\
    a3 & Linear head & $3957.4 \pm 17.8$ \\
    a4 & Random selection & $4336.8 \pm 18.5$ \\
    s1 & Active budget 500 & $4237.0 \pm 55.4$ \\
    s2 & Active budget 4000 & $4309.4 \pm 15.3$ \\
    s3 & Retrieval 32 & $3516.5 \pm 25.3$ \\
    s4 & Retrieval 128 & $5285.7 \pm 40.7$ \\
    \bottomrule
  \end{tabular}
\end{table}